\documentclass[conference]{IEEEtran}
\usepackage{amsmath}
\usepackage{amssymb}
\usepackage{graphicx}
\usepackage{subcaption}
\usepackage[ruled,vlined]{algorithm2e}
\usepackage{tabularx}
\usepackage[backend=bibtex,style=numeric-comp,sorting=none]{biblatex}
\usepackage{xcolor}
\bibliography{b} 
\usepackage{float}
\graphicspath{{figures/}}
\begin{document}

\title{A GIS Data Realistic Road Generation Approach for Traffic Simulation  }

\author{
	\IEEEauthorblockN{Yacine Amara, Abdenour Amamra, Yasmine Daheur, and Lamia Saichi}
	\IEEEauthorblockA{Ecole Militaire Polytechnique, Bordj El-Bahri BP 17, Algiers, Algeria}
	\IEEEauthorblockA{Email: amara.yacine@gmail.com}
}

\maketitle

\begin{abstract}
Road networks exist in the form of polylines with attributes within the GIS databases. Such a representation renders the geographic data impracticable for 3D road traffic simulation. In this work, we propose a method to transform raw GIS data into a realistic, operational model for real-time road traffic simulation. For instance, the proposed raw to simulation-ready data transformation is achieved through several curvature estimation, interpolation/approximation, and clustering schemes. The obtained results show the performance of our approach and prove its adequacy to real traffic simulation scenario as can be seen in this video\footnote{https://youtu.be/t8eyphcFYHc}.  
\end{abstract}

\IEEEpeerreviewmaketitle

\section{Introduction}
In recent years, applications of road traffic simulation have become ubiquitous in everyday life: driving simulation or racing games are increasingly attracting the attention of developers. However, the results obtained are not always consistent with reality. When one wants to reproduce a realistic behavior, the developer must consider the real parameters of the road.

The realization of a real-life road simulation would make it possible to forecast the traffic generated in a given road network. This realism, modeled on a computer, would help managers to detect the problems present in the network, namely congestion, accidents, user stress, an insufficient number of channels, etc. Besides optimizing traffic on the roads without making urban changes, such as the location of traffic lights, number, and width of lanes.

The shape of the road is complex, particularly at mountainous segments, severe turns, etc. Therefore, finding the best mathematical function that will fit the shape of these sections while taking into account compressing the number of data to be stored is challenging.

Geometric processing applications rely on the geometric properties of curves such as torsion, curvature, and tangents. In our study, we chose the curvature metric in the road's modeling surface, since the latter embeds the information on the shape of the road. The actual data is extracted from the Geographic Information System (GIS), the latter known for providing the ability to manipulate geographical information laid out on multiple layers. The metric being chosen; our approach is then organized as follows: extraction of the road layer as polylines in the GIS data; curvature estimation at the neighborhood of the points; a grouping of points in clusters according to the sign of their curvature, and finally the approximation of clusters by continuous mathematical functions. 

\begin{figure}[h]
	\centering   
	\includegraphics[height=3.0cm,width=0.3\textwidth]{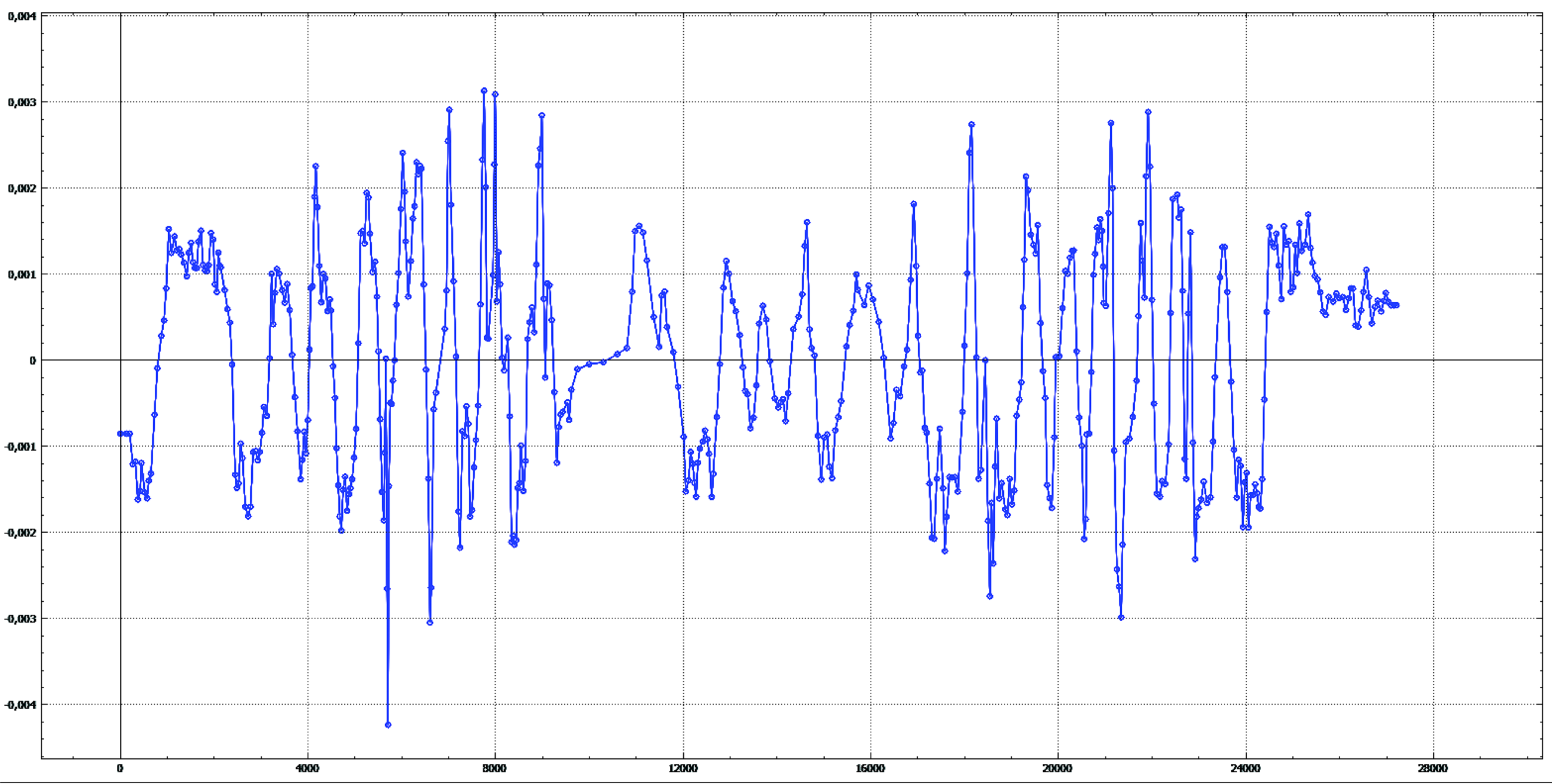}
	\caption{}
	\label{Fig_1:curvature appearance}
\end{figure}

\section{Related works}
Most of the work that has been proposed on road reconstruction is based on procedural modeling. Roads are generated by empirical rules often based on observation. We will classify the methods proposed in two categories: the methods taking the input of calculated roads (procedural approaches) and the models taking as input the data resulting from Geographic Information Systems (GIS).

\subsection{Procedural approaches}
These approaches take as input roads derived or calculated by mathematical or optimization methods. Both, Parish et al. \cite{parish2001procedural} and Jing et al. \cite{sun2002template} generated models of cities in which road networks played a key role. Their roads were built from the central lines that are generated either by grammar rules named \emph{L-Systems} \cite{parish2001procedural}, or model-based methods \cite{sun2002template}. However, the defining laws or models to regenerate exactly the structure of urban roads are not available. Hence, their roads often drift from their respective real counterparts. In addition, their methods were not designed for highways and suburban roads.

Some of these limitations have been resolved by Chen et al. \cite{chen2008interactive}, where they used a tensor field from which a graph representing the road network was calculated. This can be modified to control the locally generated road profile. Their method permits creating large areas but does not allow the reconstruction of existing networks because of the problem of scale limitation. In addition, only urban roads were considered.

Another procedural model was proposed by Galin et al. \cite{galin2011authoring, galin2010procedural} based on a platform for generating mapped roads that contain many types of features such as trees, rivers, and lakes. Their roads resulted from a short path algorithm instead of the real data. As a result, we obtain fictitious roads that do not satisfy the constraints of civil engineering. The generated roads are erroneous and do not match the shape of the terrain. This lack of realism is because of the shortcomings of the proposed approaches. The major shortcoming of these works is not to have considered the actual data as input of the problem.

\subsection{GIS data approaches}
To overcome the problems mentioned and to build more realistic roads, Bruneton and Neyret \cite{bruneton2008real} proposed a method of generating large road surfaces from GIS data. They represented the roads with Bezier curves to join the sampled points and then mapped them on the ground. Although this model helps to perceive actual road coordinates, current roads are no longer built from Bezier curves. 
Again, these models are affected by the lack of realism. The details and constraints related to civil engineering discipline are not taken into account. Road networks must be defined by simple forms such as straight lines, arcs of circles, and clothoids. This drawback had been addressed by the following work:
\subsubsection {LSGA algorithm}
This work \cite{nguyen2014realistic} introduces a new approach to construct smoothed curve pieces representing realistic roads. 
Given a GIS database of road networks, where the sampled points are organized as 3D polylines, this method creates horizontal and verticals curves, then it combines them to generate the roads. The major contribution of this work is a tree traversal algorithm that extends the sequences of the best fit primitives and a fusion process of these primitives. The latter must respect a certain grammar according to an automaton. 
This approach offers more realistic results than those that preceded it, and the errors are proportional to the noisiness of input data.

\subsubsection{Construction by clothoids}
In \cite{bruneton2008real}, the algorithm for adjusting a sequence of $G_2$ polylines into clothoid segments takes place in two steps: first a piecewise linear approximation is applied, then a sequence of rigid 2D transformations is applied in order to align the in one consistent result. Although this method respects civil engineering constraints and models the transition between a circular arc and a straight line with a clothoid, its first pass through the linear segments loses the precision when estimating the radius of curvature.

\subsubsection{Automatic generation of 3D roads}
This method, presented in \cite{Wang_2014}, is based on a set of civil engineering rules. It proposes a new approach for the automatic 3D generation of high-fidelity roads. It transforms GIS data that only contains 2D information from the central axis of the road into a 3D model of the road network. In the proposed approach, basic road elements such as road segments, road intersection are generated automatically to form sophisticated road networks. But in the modeling of the axe of the road, the segments were connected by Hermite curves, which satisfy only $G_1$ continuity, hence the civil engineering constraints were not respected.

The proposed model must be realistic, that is to say, that it must meet the constraints of civil engineering and vehicle dynamics, for this reason, our study was based on an essential criterion which is \emph{curvature}. Curvature, the inverse of the radius of the circle tangent to the curve, is defined as the norm of the acceleration vector of a body traveling the curve at unit speed. It is the second derivative with respect to the curvilinear abscissa of the body position.

\section{Curvature Estimation}

\subsection{Definition}
\newcommand{\norm}[1]{\left\lVert#1\right\rVert}
\renewcommand{\angle}{\measuredangle}
A parametric curve is a function $ r: I \subset \mathbb {R} \rightarrow \mathbb{R}^n $, when $n = 2$ it is called a \emph{plane curve}.
The curvilinear abscissa $s$ from a point $r(t_0)$, $t_0 \in I$, at a given point $r(t_1)$, $t_1 \in I$, is defined by:
\begin{equation}
	s(t_1) = \int_{t_0}^{t_1} \norm{r'(u)} du
\end{equation}
 
The vector $T(s) = r'(s)$ is called the tangent vector. The normal vector $N(s)$ is obtained by a rotation of $90^\circ$ anticlockwise. The vectors $T'(s)$ and $N(s)$ are collinear. That is, there is a function $k(s)$ such that:
\begin{equation}
	T'(s) = k(s) \times N (s)
\end{equation}
called the \emph{curvature} of the curve at the point $r(s)$. The curvature also corresponds to the variation of the direction of the tangent vector respectively to the curvilinear abscissa: $k(s) = \theta'(s)$, such that :
\begin{equation}
		\theta'(s) = \angle (\overrightarrow{T(s)},\overrightarrow{(1,0)})
\end{equation}
Since our initial data is in discrete form, we performed a local estimation using a sliding window. The latter is centered around a point allowing to approximate all it neighbors within the window by a second order polynomial. The interest of such an operation is to deduce the first and the second derivatives, to then calculate the curvature. Several approximation methods called \emph{implicit parabola fitting} proposed in \cite{Lewiner_2005}, which have a good performance and a fair simplicity of implementation.

\subsection{Second-order curve approximation}
In the implicit parabola fitting method, proposed in \cite{Lewiner_2005}, the curve is described by a function such that: $y=f(x)$ or $x=f(y)$. The variation of $x$ and $y$ inside the window determines the parameterization to adopt, in fact, if the variation of $x$ is greater than that of $y$, the algorithm will select the case $y=f(x)$ and vice versa .
In order to simplify the notation, we will consider that $p_0 = (0,0)$. The goal is to find $f_{0}^{'}$ and $f_{0}^{''}$, which minimize:
\begin{equation}
		E_x(f_{0}^{'} , f_{0}^{''})  = \sum_{i=-q}^{q} (y_i - f_{0}^{'} x_i - \frac{1}{2} f_{0}^{''} x_i^2 )^2
\end{equation}

The solution of this problem of least squares gives:

\begin{equation}
	f_{0}^{'} = \frac{cg-bh}{ac-b^2} \quad f_{0}^{''} = \frac{ah-bg}{ac-b^2}
\end{equation}
such that : \\
\begin{equation}
	\begin{split}
		a = \sum_{i=-q}^{q} x_i^2\quad
		g= \sum_{i=-q}^{q} x_i y_i\quad 
		b = \frac{1}{2} \sum_{i=-q}^{q} x_i^3,\\
		h = \frac{1}{2} \sum_{i=-q}^{q} x_i^2  y_i\quad
		c = \frac{1}{4} \sum_{i=-q}^{q} x_i^4
	\end{split}
\end{equation}

\begin{figure}[h]
	\centering   
	\begin{subfigure}[b]{0.3\textwidth}
		\includegraphics[height=3.0cm,width=\textwidth]{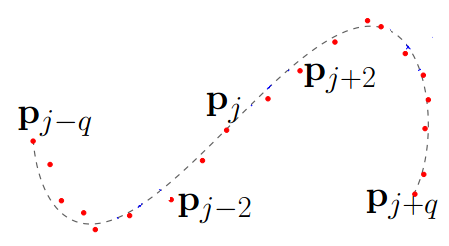}
		\caption{}
		\label{Fig_5:Point window}
	\end{subfigure}
	\hfill
	\begin{subfigure}[b]{0.3\textwidth}
		\includegraphics[height=3.0cm, width=\textwidth]{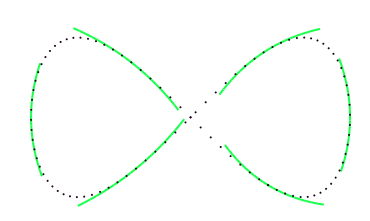}
		\caption{}
		\label{Fig_5:Approximation by least squares}
	\end{subfigure}
	\caption{(a) Points window, (b) Approximation by least squares}
	\label{Fig_5}
\end{figure}

\subsection{Curvature computation}
\subsubsection{Model and notation}
Consider a set of points ${p_i}$ of a flat smooth curve $r$, in this study the curve is parameterized by the arc length, the estimate of the curvature for a plane curve requires an approximation of the first and second derivatives of ${r}(s)$. 
Let a point $p_0$ be chosen  . The derivation of $r$ in $p_0$ will be estimated from a window of size $2q + 1$ around $p_0 : {p_{-q}, p_{-q + 1}, \ldots, p_q}$ (see Fig \ref{Fig_5:Point window}). We set $p_{0} = r(0)$ as the origin, the approximation of degree two can be written in the form:
\begin{equation}
	r(s) = r'(0)s+\frac{1}{2}+r''(0)s^2
\end{equation}

\subsubsection{The least squares approach}
The estimate of $r'(0)$, $r''(0)$ is obtained by the least squares approach (Fig \ref{Fig_5:Approximation by least squares}). The weighting $w_i$ of the point $p_i$ must be positive, relatively important for small values of $|s_i|$  and relatively small for large values of $|s_i|$.

The arc length $s_i$ can be estimated as follows: $\Delta l_k$ is the arc length of the vector $p_k p_{k+1}$, where $k$ varies from $-q$ to $q-1$. The arc length between $p_0$ and $p_i$ can be approximated by:

\begin{equation}
\begin{aligned}
	&\left\{
	\begin{array}{ll}
		l_i = \sum_{k=0}^{i-1} , i>0 \\ 
		l_i = -\sum_{k=0}^{-1} , i<0
	\end{array}
	\right.
	\end{aligned}
\end{equation}

\subsection{Curvature computation for a plane curve}
For the case of plane curves, the idea of the Independent coordinates method \cite{Lewiner_2005} is to construct a parametric curve $(x(s), y(s))$ that approaches the curve locally, by quadratic functions as a function of arc length.
\begin{equation}
	\begin{aligned}
	&\left\{
	\begin{array}{ll}
		\hat{x}(s)=x_0+x_{0}^{\prime}s+\frac{1}{2} x_0^{\prime\prime}s^2 \\
		\hat{y}(s)=y_0+y_{0}^{\prime}s+\frac{1}{2} y_0^{\prime\prime}s^2
	\end{array}
	\right.
	\end{aligned}
\end{equation}
The derivatives $x_{0}^{\prime}$ and $x_{0}^{\prime\prime}$ are estimated by minimizing:

\begin{equation}
	E_x (x_{0}^{\prime}, x_{0}^{\prime \prime}) = \sum_{i=-q}^{q} w_i (x_i - x_{i}^{\prime} l_i - \frac{1}{2} x_{0}^{\prime\prime} l_{i}^{2} )^2
\end{equation}
The minimization of this equation can be written in the following matrix form
\begin{equation}
	\begin{bmatrix}
		a_{1}       & a_{2}  \\
		a_{2}       & a_{3} 
	\end{bmatrix}
	\begin{bmatrix}
		x_{0}^{\prime} \\
		x_{0}^{\prime \prime} \\
	\end{bmatrix}
	=
	\begin{bmatrix}
		b_{x,1} \\
		b_{x,1} \\
	\end{bmatrix}
\end{equation}
such that : \\
\begin{equation}
	\begin{split}
		a_1 = \sum_{i=-q}^{q} w_i l_{i}^{2}\quad 
		a_2 =  \frac{1}{2} \sum_{i=-q}^{q} w_i l_{i}^{3}\quad
		a_3 =  \frac{1}{4} \sum_{i=-q}^{q} w_i l_{i}^{4}  \\
		b_{x,1} = \sum_{i=-q}^{q} x_i w_i l_{i}\quad
		b_{x,2} = \frac{1}{2} \sum_{i=-q}^{q} w_i l_{i}^{2} x_i
	\end{split}
\end{equation}

\begin{algorithm}
	\SetAlgoLined
	\DontPrintSemicolon
	$I[ ]=a_1=a_2=a_3=b_{x,1}=b_{x,2}=b_{y,1}=b_{y,2}=0$ \;
	\For{i = -q; i $\leq q$; q++}
	{
		$I[i] \gets I[i-1] + \norm{p_{i}p_{i-1}}$\;
	}
	$m = I[0]$\;
	\For{i = -q; i $leq q$; q++}
	{
		$I[i] \gets I[i] - m$\;
		$w \gets weight(I[i])^2$\;
		$a_1 \gets a_1 + w(I[i])^2$\;
		$a_2 \gets a_2 + \frac{w}{2}(I[i])^3$\;
		$a_2 \gets a_2 + \frac{w}{4}(I[i])^4$\;
		$b_{x,1} \gets b_{x,1} + w(I[i])(x_i - x_0)$\;
		$b_{y,1} \gets b_{y,1} + w(I[i])(y_i - y_0)$\;
		$b_{x,2} \gets b_{x,1} + \frac{w}{2} (I[i])^2(x_i - x_0)$\;
		$b_{y,2} \gets b_{y,1} + \frac{w}{2} (I[i])^2(y_i - y_0)$\;
	}
	$d \gets a_1 a_3 - a_{2}^{2}$\;
	\caption{Weighted least square variables setting}
\end{algorithm}

\begin{algorithm}
	\SetAlgoLined
	\DontPrintSemicolon
	$x_{0}^{\prime} \gets (a_3 b_{x,1} - a_2 b_{x,2})/d$ \;
	$y_{0}^{\prime} \gets (a_3 b_{y,1} - a_2 b_{y,2})/d$ \;
	$x_{0}^{\prime\prime} \gets (a_1 b_{x,2} - a_2 b_{x,1})/d$ \;
	$y_{0}^{\prime\prime} \gets (a_1 b_{y,2} - a_2 b_{y,1})/d$ \;
	$\kappa \gets (x_{0}^{\prime} y_{0}^{\prime\prime} - y_{0}^{\prime} x_{0}^{\prime\prime}) / \norm{(x_{0}^{\prime}, y_{0}^{\prime})}^3$\;
	$T \gets (x_{0}^{\prime}, y_{0}^{\prime}) / \norm{(x_{0}^{\prime}, y_{0}^{\prime})}$\;
	$N \gets sign(\kappa)(-T_{y} , T_{x})$\;
	\caption{Coefficient computation}
\end{algorithm}

The same procedure is applied for the calculation of $y_{0}^{\prime}$ and $y_{0}^{\prime \prime}$.
The tangent $T$ is obtained by the normalization of the vector $r_{0}^{\prime}= (x_{0}^{\prime}, y_{0}^{\prime})$, while the normal vector is obtained by a rotation of $90^\circ$ of $T$.

\section{Cutting according to curvature}
This step will aim to cut the road into a set of primitives consisting of straight lines, left/right turns based on the value of curvature estimated at each point as explained in the previous step.

For this, we will execute the following processes in the order indicated. We first start with a classification in two primitives only, that is to say left/right turn, we continue by extracting the sections which are straight lines independently of the primitives obtained previously and we thus finish by eliminating the isolated points, which are of the right type because a primitive of the right type must have a minimum number of two points.
In what follows we will detail the procedure for filtering the curvature values according to the order of execution of the steps.

\subsection{Preprocessing}
In order to facilitate the different treatments, an initial marking of all the points of the curve has been carried out. For this, we assigned to each point whose curvature is positive the number +1, and -1 for those whose value of the curvature is negative, and we stored the result in a table named "ids" which will be subsequently updated by the results of the different processing.

\subsection{Left and right turns detection}
Our approach was based on a local estimate of the curvature sign near the point considered. A sliding window of size $w$ has been used to estimate the global sign of the values inside the window by summing the values previously assigned, so the classification of the types of sections will be carried out as following :
\begin{itemize}
	\item The sum is positive, so it is a right turn and the point is marked +1.
	\item The sum is negative, so it is a left turn and the point is marked -1.
	\item The zero sum is not a possible case because the size of the window around the point is odd.
\end{itemize}

This step is very important because it allows to assign a point to a given turn even if the sign of its curvature does not correspond, because it is the trend of the whole neighborhood is taken into consideration.

\subsection{Straight line detection}

Straight line detection is based on the value of the estimated curvature. Indeed, we know that a line is characterized by a null curvature. Since the values of the estimated curvature are not all accurate and contain noise, a threshold $\delta$ has been set. If the curvature norm is below this threshold, the point will be assigned to a straight line primitive. Points satisfying the relationship $|K(s)| < \delta$  will be assigned the number 0.

\subsection{Elimination of isolated right point}
Knowing that a marked point of a straight line can not be isolated, the solution is to go through all the points of the line and to modify those which are of marking different from their neighbors according to the signs of the curvature at this point, i.e. +1 marking if the curvature is positive, -1 otherwise.
\begin{algorithm}
	\caption{Primitive assignment}
	\label{alg:Primitive assignment}
	\SetAlgoLined
	\DontPrintSemicolon
	\While{i $<$ ids.size()}{
		gpts gpt \;
		gpt.type $\gets$ ids[i] \;
		gpt.id $\gets$ i \;
		gpt.nbr $\gets$ 0 \;
		\For{i = 0; i $<$ ids.size(); i++}{
			\If{gpt.type $\neq$ ids[i]}  
			{
				Break \;
			} 
			gpt.nbr++ \;	
		}
	}
\end{algorithm}
\section{Turns approximation with second order polynomials}
\subsection{Preprocessing}
In order to interpolate curves with polynomials, a data structure has been created to facilitate processing on the one hand and to save the results obtained on the other hand. This data structure ("gpts") consists of an integer field named "type", which will contain the type of the primitive according to the marking carried out in the previous step. An integer field named "id", which will contain the identifier of the first point of the primitive. An integer field named "nbr", which will contain the number of points belonging to this primitive as well as two other fields "paramX" and "paramY", which will be used to store the coefficients of polynomials associated with turns.
After the structure is created, the fields were subsequently assigned according to algorithm ~\ref{alg:Primitive assignment}.

For each left-handed or right-turn type primitive, we will proceed to the parameterization according to the length of the arc, so we will have two polynomials according to $x$ and $y$ such that:

\begin{equation}
	\begin{aligned}
	&\left\{
	\begin{array}{ll}
		x(s)=a_0+a_{1}s+a_{2}s^2+a_{3}s^3 \\
		y(s)=b_0+b_{1}s+b_{2}s^2+b_{3}s^3
	\end{array}
	\right.
	\end{aligned}
\end{equation}
The polynomials in question will be of degree 3, this choice can be justified by the absence of inflection points since the primitives have been classified according to their (there is not a transition in the same road segment). They are convex or concave curves hence the choice of the polynomial regression that is a statistical analysis that describes the variation of an dependent random variable, called here $x$ or $y$, according to an independent random variable, called here $s$, being the length of the arc. We seek, by regression, to bind the variables by a polynomial of degree 3.
The calculation of the coefficients therefore amounts to solving a system of equations that can be expressed in the following matrix form:
\begin{equation}
	\begin{pmatrix}
	 x_{1} \\
	 x_{2} \\
	\vdots \\
	x_{n} 
	\end{pmatrix}
	 = 
	\begin{pmatrix}
	1 & s_{1}^{1} & s_{1}^{2} \cdots & s_{1}^{m}  \\
	1 & s_{2}^{1} & s_{2}^{2} \cdots & s_{2}^{m}  \\
	\vdots  & \vdots  & \ddots & \vdots  \\
	1 & s_{n}^{1} & s_{n}^{2} \cdots & s_{n}^{m}  \\ 
	\end{pmatrix}
	\begin{pmatrix}
	a_{0} \\
	a_{1} \\
	\vdots \\
	a_{m} 
	\end{pmatrix}
\end{equation}
The number $m = 3$ corresponds, in our case, to the cubic polynomial regression. We will treat the case of the polynomial $x(s)$ generated and it is the same for $y(s)$. The problem is therefore to find the vector $\vec{a}$ in the equation:
\begin{equation}
  \vec{x} = S \vec{a}
\end{equation}
The solution is given by:
\begin{equation}
	\vec{a} = (S^{T}S)^{-1} S^{T}\vec{x}
\end{equation}
\subsection{Error criterion}
Given the needs and requirements of the realism imposed on the generated road, we set ourselves the objective of approximating the GIS input data with lines and polynomials of third degrees such that the maximum deviation between the two curves does not exceed 2.0m, in this part we discuss the calculation of error as well as treatments to be undertaken in the case of exceeding this limit.

\subsubsection{Error computation}
The error $d$ is estimated as the distance between the approximate point $p=(x, y)$ and the entry point $p=(x, y)$, it is deduced according to the following formula:
\begin{equation}
	d = \sqrt{(\hat{x}(s) - x(s))^2 + (\hat{y}(s) - y(s))^2 }
\end{equation}
The maximum value of this error must not exceed $\delta$ for each of the points of the different right or polynomial primitives.
\begin{equation}
	d_{max} <\delta 
\end{equation}
\subsubsection{Error processing}
Primitives whose error exceeds the required threshold must be modified, for this our approach is to generate from the initial primitive two primitives such that the number of points constituting each primitive is equal to half the number of points of the primitive. For a given primitive $C_m$ one generates two primitives $C_{m/2}$ and $ C_{m/2}^{\prime}$, and the same treatment will be executed on the two new primitives in a recursive way, this will assure us:
\begin{itemize}
	\item Compliance with the error limit required for each final primitive obtained,
	\item The generation of a minimal number of primitives,
\end{itemize}

\section{Results}
The results obtained when applying our method are summarized in Figure (\ref{Fig_6}). The figures show the stages of our approach. Figure (\ref{Fig_6_1}) shows a sample of a road polyline. In Figure (\ref{Fig_6_2}), the curvatures were estimated by the least squares method. Note that this representation is more exploitable than that of Figure (\ref{Fig_1:curvature appearance}), and that there is less disturbance in the values of the curvatures due to windowing.

\subsection{Curvature estimation by the implicit parabola fitting method}
In order to overcome the problems encountered when calculating the curvature by cubic splines, we opted for an estimation of the latter by the least squares. Compared with spline, the quality of curvature values improved significantly. Indeed, most points follow the trend of the turn where they belong as can be seen in Figure (\ref{Fig_6_2}). Nevertheless, in the case of more complex turns, outliers appear. Hence, we have proposed to neglect them, and to mark a window by a single sign of curvature relative to reality.

\subsection{Clustering}
\begin{figure}[h]
	\centering   
	\begin{subfigure}[b]{0.48\linewidth}
		\includegraphics[height=2.5cm, width=\linewidth]{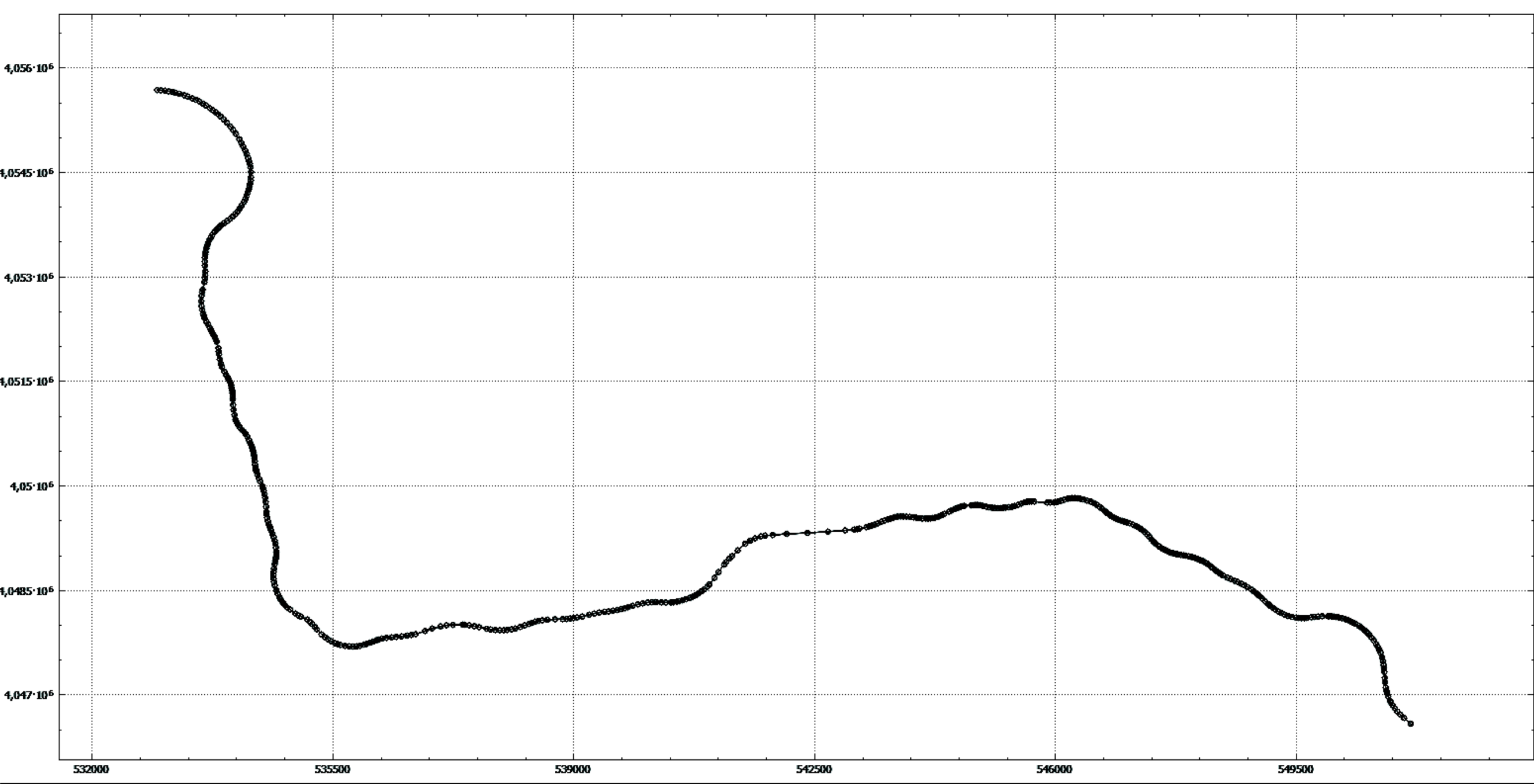}
		\caption{}
		\label{Fig_6_1}
	\end{subfigure}
	\hspace{0.1cm}
	\begin{subfigure}[b]{0.48\linewidth}
		\includegraphics[height=2.5cm, width=\linewidth]{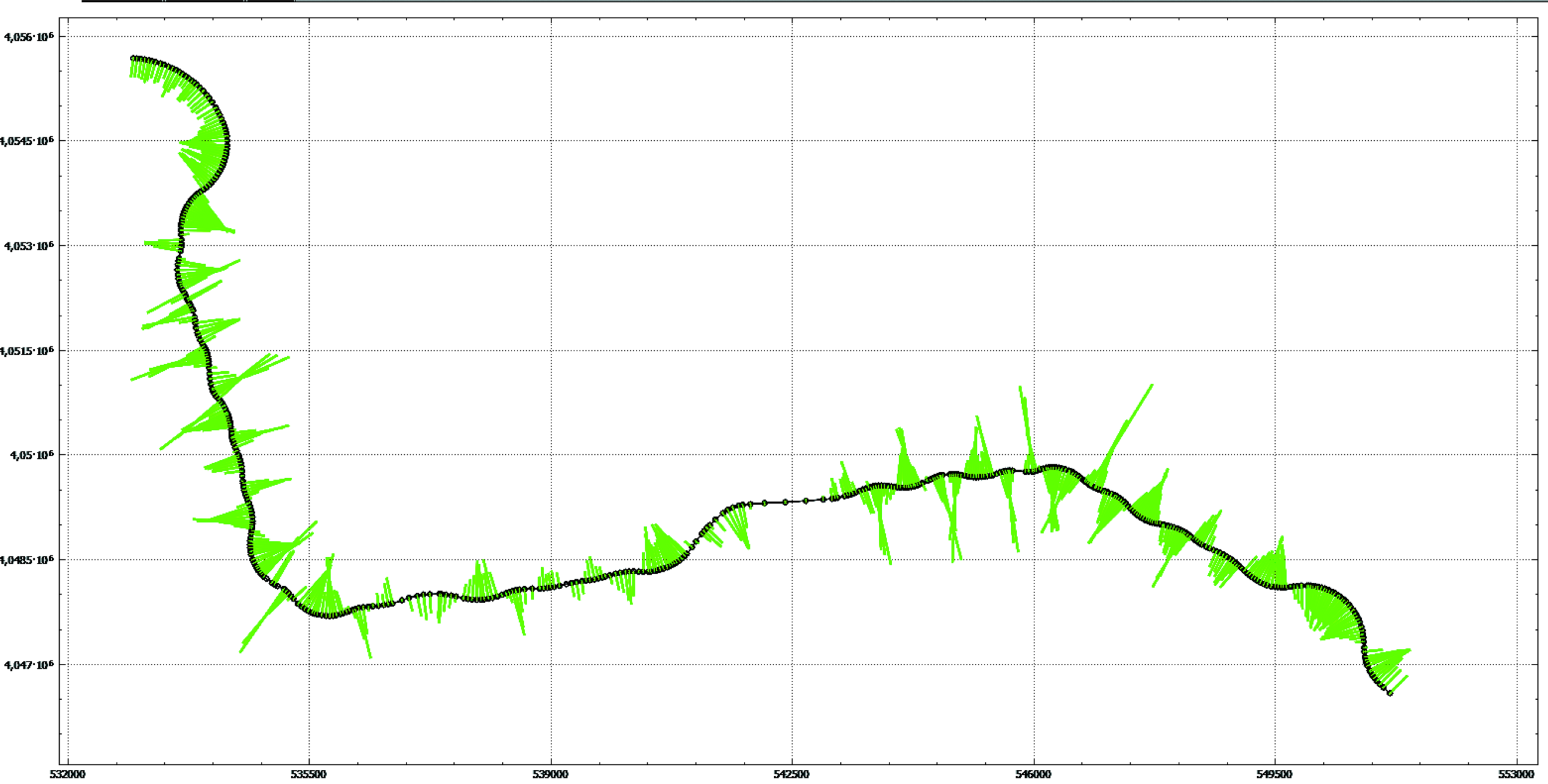}
		\caption{}
		\label{Fig_6_2}
	\end{subfigure}

	\begin{subfigure}[b]{0.48\linewidth}
	\includegraphics[height=2.5cm, width=\linewidth]{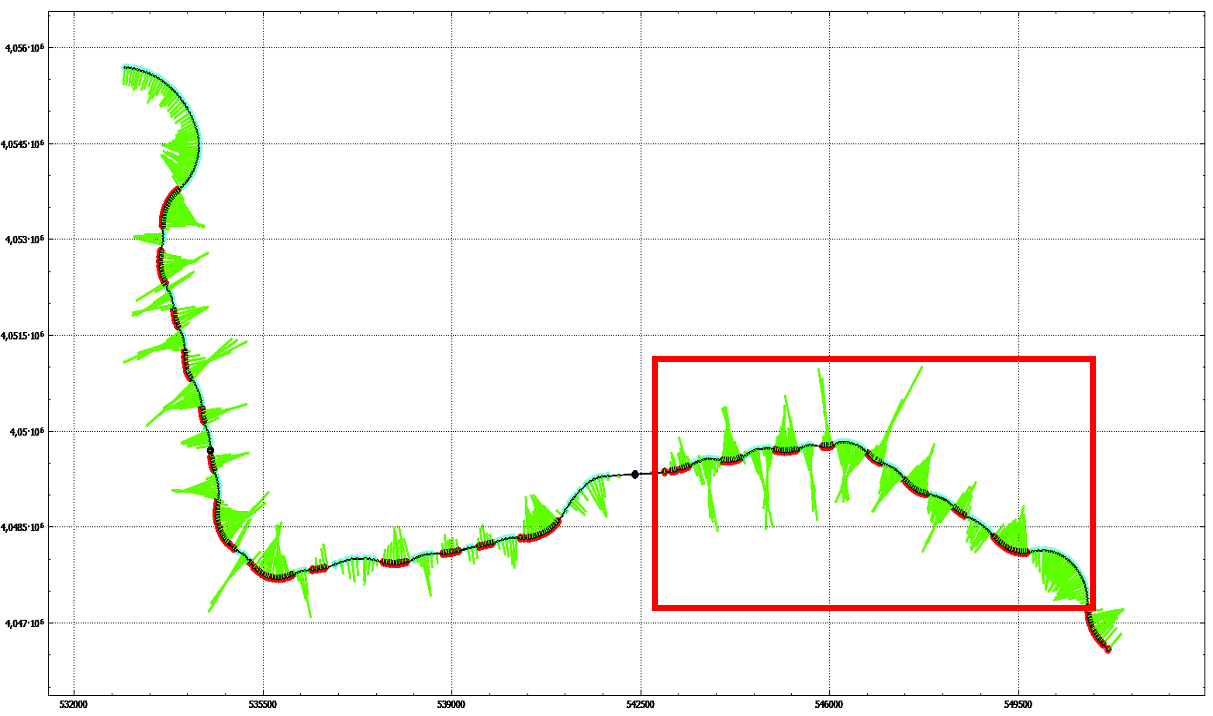}
	\caption{}
	\label{Fig_6_3}
	\end{subfigure}
	\hspace{0.1cm}
	\begin{subfigure}[b]{0.48\linewidth}
		\includegraphics[height=2.5cm, width=\linewidth]{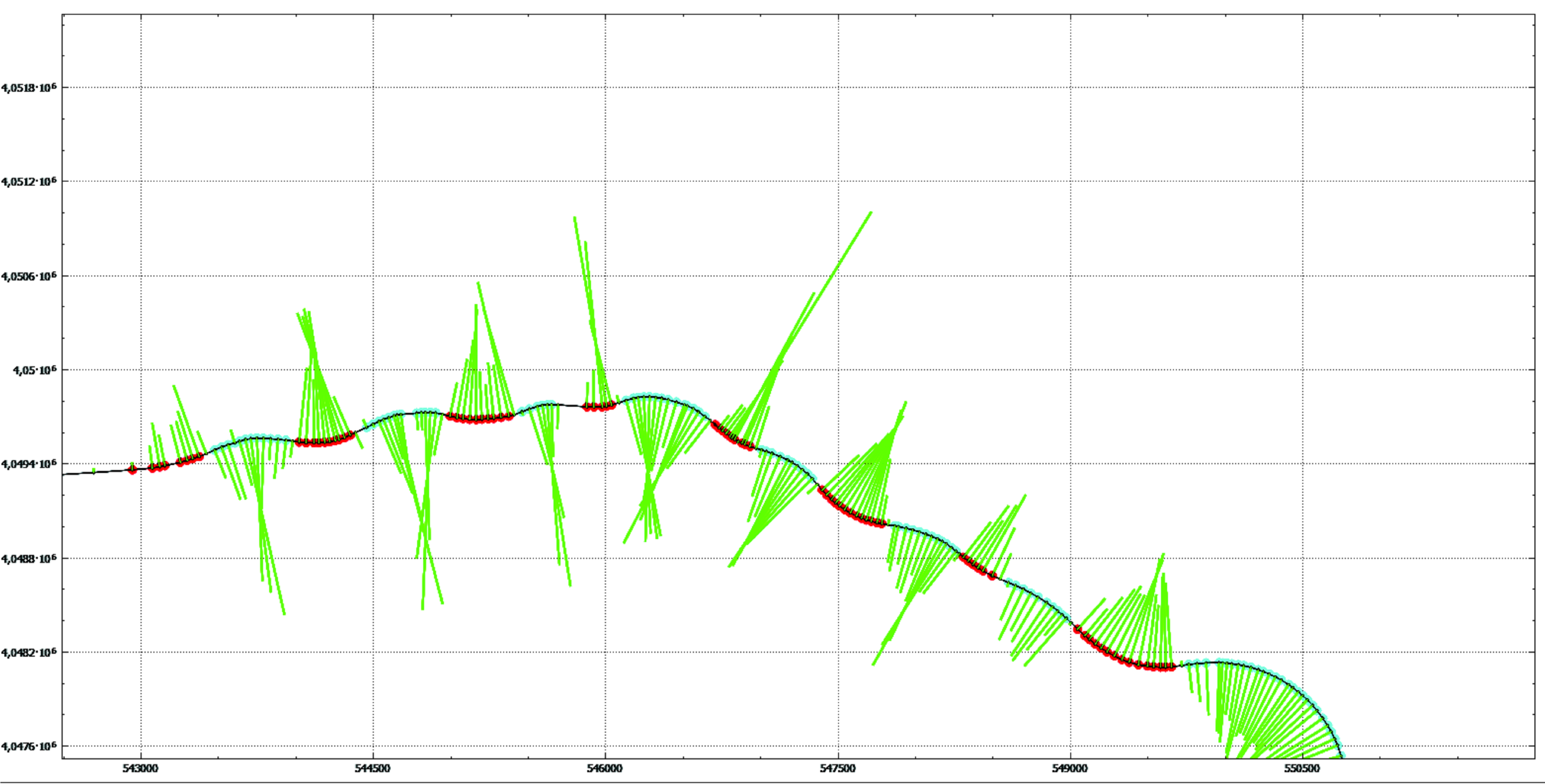}
		\caption{}
		\label{Fig_6_4}
	\end{subfigure}

	\begin{subfigure}[b]{0.48\linewidth}
	\includegraphics[height=2.5cm, width=\linewidth]{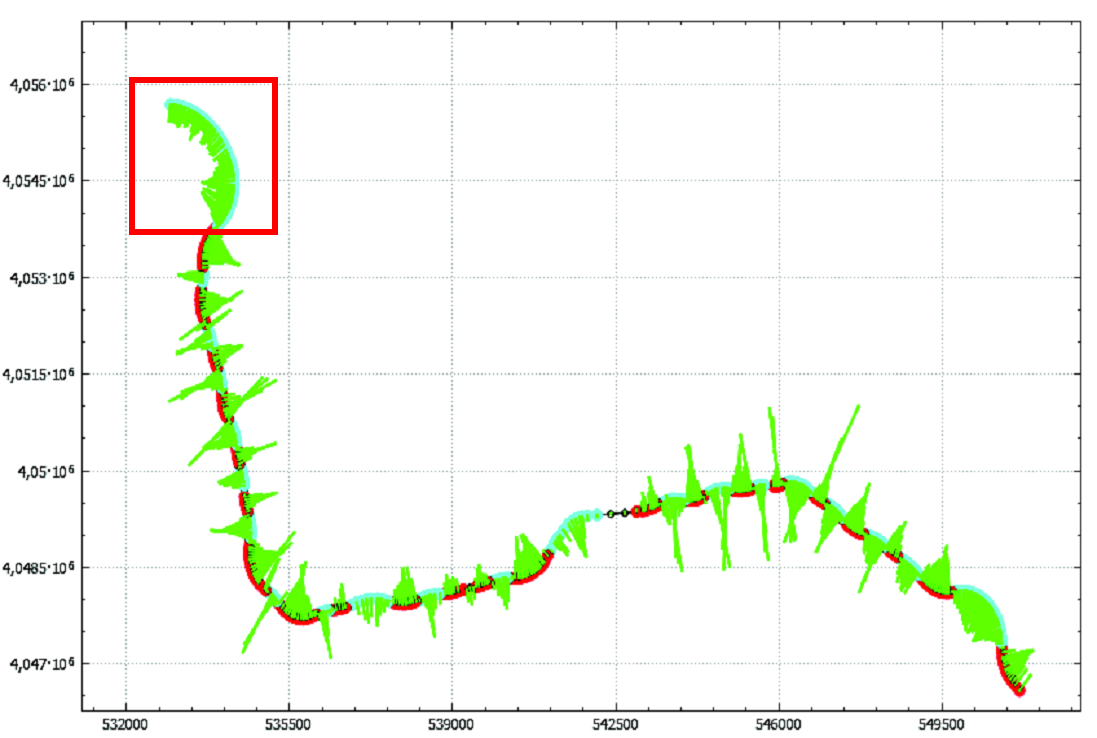}
	\caption{}
	\label{Fig_6_5}
	\end{subfigure}
	\hspace{0.1cm}
	\begin{subfigure}[b]{0.48\linewidth}
		\includegraphics[height=2.5cm, width=\linewidth]{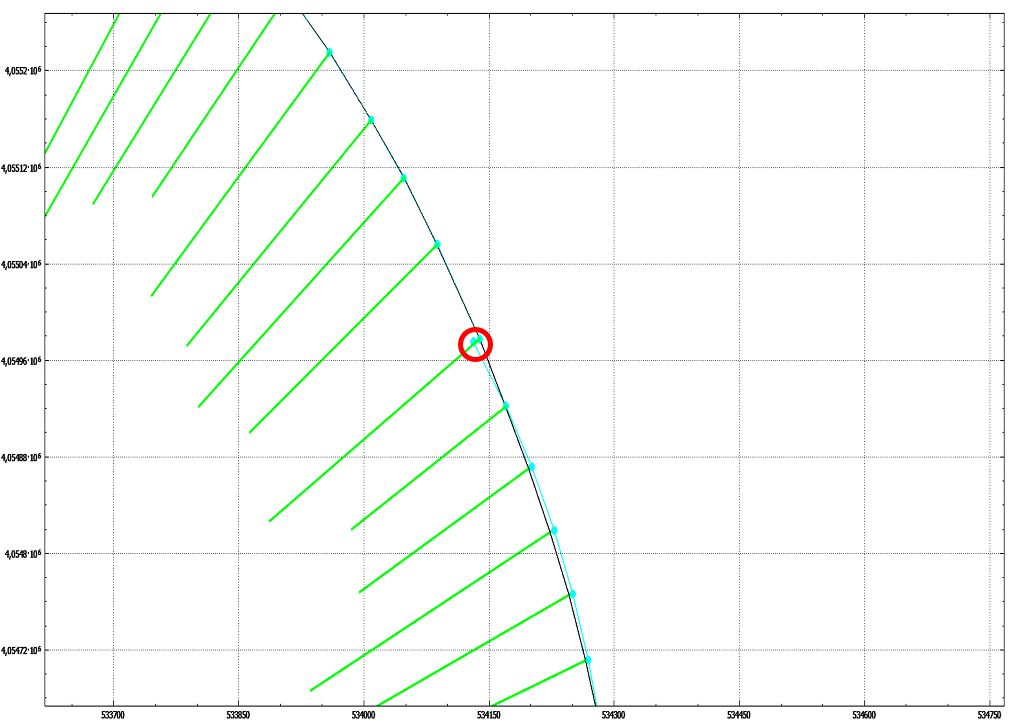}
		\caption{}
		\label{Fig_6_6}
	\end{subfigure}
	\begin{subfigure}[b]{0.48\linewidth}
	\includegraphics[height=2.5cm, width=\linewidth]{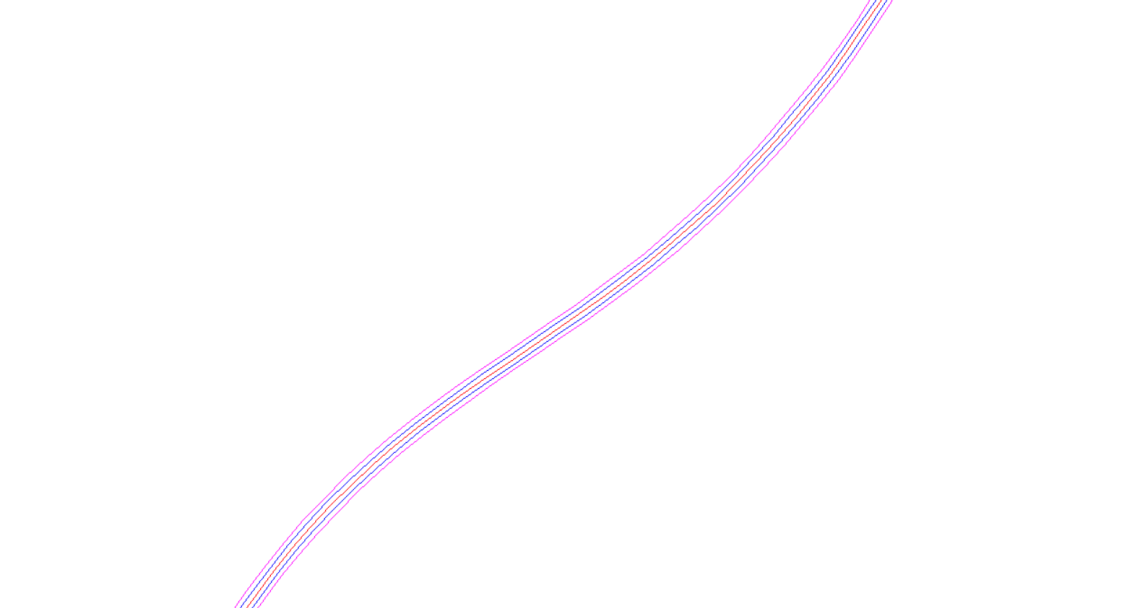}
	\caption{}
	\label{Fig_6_7}
	\end{subfigure}
	\hspace{0.1cm}
	\begin{subfigure}[b]{0.48\linewidth}
		\includegraphics[height=2.5cm, width=\linewidth]{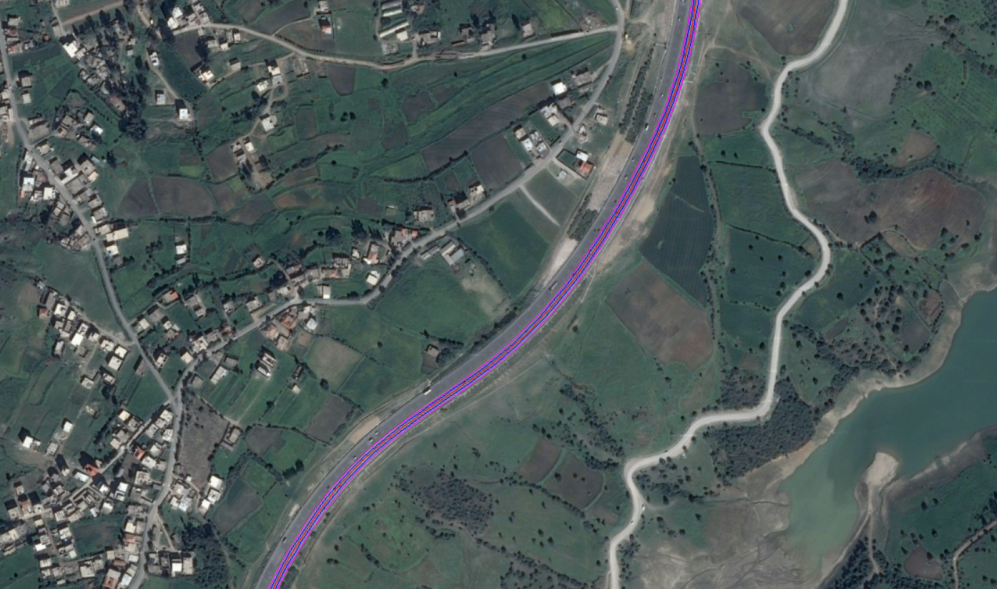}
		\caption{}
		\label{Fig_6_8}
	\end{subfigure}
	\caption{Processing steps of the proposed algorithms. 
		(a) Raw road in polyline form,
		(b) Curvature obtained by least squares according to the normal,
		(c) Cutting of the road (left, right turns and straight lines) according to curvature values,
		(d) Turns detected by the cutting algorithm,
		(e) Polynomial fitting of the detected turns,
		(f) 3rd degree polynomials approximation of a given turn,
		(g) Result: generated road surface,
		(h) Mapping of the result on a real road}
	\label{Fig_6}
	\end{figure}

Figure (\ref{Fig_6_3}) shows that the points of the input road were grouped into three basic categories: left turn, right turn, and straight line. We note that the results are more refined, because this step corrects any possible residual error of the previous section. Moreover, it allows noise reduction, by defining the straight line segments from a given threshold, since the SIG data being noisy, we cannot obtain zero curvature values.

This step allowed us to have the same signs of curvature for a given type of cluster. Nevertheless, at this stage the clusters are not connected to each other, a major problem on which the approximation will be based in the following phases.

To overcome the problem of connection, we fix first the ends of the primitives, then we approximate the calculated model of the initial points. In this perspective, the Bezier curves with a least squares approximation prove to be an adequate choice. Indeed, the first and last control point are superimposed on their correspondents in the initial data, then the other points are calculated by minimizing the differences between the model and the initial data. The main disadvantage of this method is that it guaranties only $C_0$ continuity between two clusters of successive points, which does not satisfy civil engineering constraints. Moreover, in some complex turns, it remains difficult to follow the shape of the cluster by a third degree polynomial.

Figure (\ref{Fig_6_7}, \ref{Fig_6_8}) represents the constructed road. The initial data points circled and clustered. Note that the resulting model does not deviate from the input road polyline. The curves in magentas represent the edges of the road, the input data (polylines) are in red, and the model (at the central axis) is in blue. Figure (\ref{Fig_6_8}) illustrates the mapping of the road to the geo-referenced satellite image corresponding to the road section. As can be seen, the model provides a smooth representation of the road surface. We see that complex shapes such as turns are well approximated. In addition, our model meets the $C_2$ continuity (imposed by road civil engineering), along the road without oscillations or other erratic behaviors that would compromise the visual comfort during the  simulation.

\section{Conclusion and future works}

The objective of this work was the realistic modeling of the road surface by taking into account a number of civil engineering and vehicle dynamics constraints. We chose curvature as a parameter describing the shape of the road in order to approach the profile of the latter as closely as possible.

We proceed through the estimation of curvature by a windowing approach because of the discrete nature of GIS data. In order to approach the reality, we grouped the points into clusters according to their neighborhood's curvature to obtain basic forms of the road namely left/right turns and straight lines.

This being done, the next step was to find the mathematical functions as well as the appropriate conditions and constraints to approach the initial data with functions that respect $C_2$ continuity conditions. 

The results obtained are satisfactory insofar as our road reconstruction approach takes into account the real constraints, moreover the model obtained is of $C_2$ continuity, which reflects the smoothness of the position, speed and the acceleration of the simulated vehicle.

As a perspective, we plan to improve and refine this work by considering road intersections, since the latter are frequently encountered in a real-life situation.

\printbibliography
\end{document}